# Improving Iris Recognition Accuracy by Score Based Fusion Method

**Ujwalla Gawande**

ujwallgawande@yahoo.co.in

Computer Technology Department

YCCE, Nagpur, Maharashtra, India

**Mukesh Zaveri**

mazaveri@coed.svnit.ac.in

Computer Engineering Department

SVNIT, Surat, Gujarat, India

**Avichal Kapur**

avichal.kapur@yahoo.in

NYSS

Nagpur, Maharashtra, India

**Abstract**

Iris recognition technology, used to identify individuals by photographing the iris of their eye, has become popular in security applications because of its ease of use, accuracy, and safety in controlling access to high-security areas. Fusion of multiple algorithms for biometric verification performance improvement has received considerable attention. The proposed method combines the zero-crossing 1 D wavelet Euler No., and genetic algorithm based for feature extraction. The output from these three algorithms is normalized and their score are fused to decide whether the user is genuine or imposter. This new strategies is discussed in this paper, in order to compute a multimodal combined score.

**Keywords:** Iris, Score, Feature Extraction, Matching, Fusion

## 1. Introduction

In recent years, the authentication elements Biometric recognition is a common and reliable way to authenticate the identity of a living person based on physiological or behavioral characteristics. A physiological characteristic is relatively stable physical. In information technology, in particular, biometrics is used as a form of identity access management and access control. It is also used to identify individuals in groups that are under surveillance. A detailed literature survey of iris recognition algorithm can be found in [1].In verification or authentication, a claim is made concerning the identity of a person. The biometric system has to take a binary decision of accepting or rejecting an individual, based on the information extracted from the considered biometric trait. In a verification context, two situation of error are possible, an imposter is accepted or the correct user is rejected. Performance measure of verification systems is related to the frequency with which this situation of error happens. One common performance measure, for example is the equal error rate.

The iris is a thin circular diaphragm, which lies between the cornea and the lens of the human eye. The iris is perforated close to its center by a circular aperture known as the pupil. The function of the iris is to control the amount of light entering through the pupil, and this is done by the sphincter and the dilator muscles, which adjust the size of the pupil. The average diameter of the iris is 12 mm, and the pupil size can vary from 10% to 80% of the iris diameter.

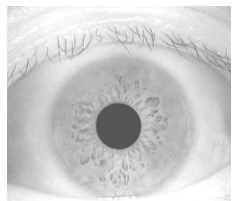

FIGURE 1: Example of Iris

An automated use of iris recognition as a means of authentication has been originally proposed by Flom and Safir. Daugman has proposed an operational iris recognition system in 1994 [2]. Since then, iris biometric is evolved as a standard reference model for verification. Iris recognition technology offers the highest accuracy in identifying individuals of any method available.

This is because no two irises are alike - not between identical twins, or even between the left and right eye of the same person. Irises are also stable; unlike other identifying characteristics that can change with age; the pattern of one's iris is fully formed by ten months of age and remains the same for the duration of their lifetime [3]. Iris recognition technology is also accurate because it uses more than 240 points of reference, in iris pattern, as a basis for a match.

Some studies [4] have showed that the performance of any single trait verification system can be improved by unimodal (or bimodal) fusion. i.e. the combination of various verification strategies applied on the same input data. Even grater verification performance improvement can be expected trough the use of multiple biometric characteristics if we assume statistical independence between them. Work related to multimodal fusion approach is given in [5] [6] [7]. Multialgorithmic biometric systems take a single sample from a single sensor and process that sample with two or more different algorithms. The technique could be applied to any modality. Maximum benefit would be derived from algorithms that are based on distinctly different and independent principles. An intelligent fusion algorithm combines the score to improve the iris recognition performance and reduce the false rejection rate [8] [9].

**1.1   Proposed Approach**

Our basic study of the Daugman's mathematical algorithms for iris processing, derived from the information found in the open literature, led us to suggest a few possible methods [2]. Iris recognition technology works by combining computer vision, pattern recognition, and optics. First, a black-and-white video camera zooms in on the iris and records a sharp image of it. The iris is lit by a low-level light to aid the camera in focusing. A frame from this video is then digitized into a 512 byte file and stored on a computer database.

There are three main stages in iris recognition system
- Image preprocessing
- Feature Extraction
- Template Matching

In this paper we use hough transform for localization and segmentation of iris image. Dougmans rubber sheet model for iris normalization. For feature extraction we are using three distinct algorithms i.e. zero-crossing 1 D wavelet Euler No., and genetic algorithm. The score basefusion approach is used to make a final decision of accepting or rejecting the user.

The paper is organized as follows: section 2 describes the image preprocessing using Hough transform and iris normalization using Dougmans rubber sheet model. Section 3 explains the feature extraction using zero-crossing 1 D wavelet Euler No., and genetic algorithm. Section 4 describes about template matching using hamming distance. Section 5 proposed the score normalization and fusion techniques. Section 6 summarizes the paper.

## 2. Image Preprocessing

The iris image needs to be preprocessed to obtain useful iris region. Image preprocessing is divided into three steps:
- Iris localization
- Iris normalization
- Image enhancement.

### *2.1. Iris Localization*

Iris localization detects the inner and outer boundaries of the iris. Both the inner and outer iris boundaries can be approximately modeled as circles. The center of iris does not necessarily concentric with the center of pupil. Iris localization is important because correct iris region is needed to generate the templates for accurate matching. The eyelids and eyelashes normally occlude the upper and lower parts of the iris region. Also, specular reflections can occur within the iris region corrupting the iris pattern. A technique is required to isolate and exclude these artifacts as well as locating the circular iris region as shown in figure 2.

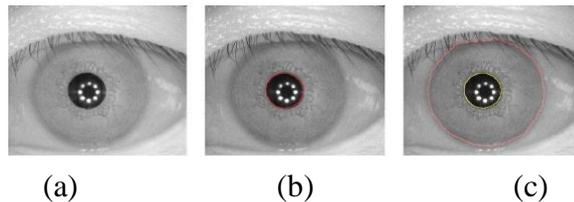

(a)   (b)   (c)
FIGURE 2:  (a) Original image. (b) Pupil boundary. (c) final iris and pupil boundary.

We use Hough transform for localization and segmentation of the iris.

### *2.1.1 Hough Transform*

The Hough transform is a standard computer vision algorithm that can be used to determine the parameters of simple geometric objects, such as lines and circles, present in an image. The circular Hough transform can be employed to deduce the radius and center coordinates of the pupil and iris regions. An automatic segmentation algorithm based on the circular Hough transform is employed by Wildes et al. [10], Kong and Zhang [11]. Firstly, an edge map is generated by calculating the first derivaties of intensity values in an eye image and then thresholding the result. From the edge map, votes are cast in Hough space for the parameters of circles passing through each edge point. These parameters are the center coordinates $x_c$ and $y_c$, and the radius $r$, which are able to define any circle according to the equation $x_c^2 + y_c^2 - r^2$

A maximum point in the Hough space will correspond to the radius and center coordinates of the circle best defined by the edge points. Wildes et al. and Kong and Zhang also make use of the parabolic Hough transform to detect the eyelids, approximating the upper and lower eyelids with parabolic arcs, which are represented as;

$$(-(x-h_j)\sin\theta_j + (y-k_j)\cos\theta_j)^2 = a_j((x-h_j)\cos\theta_j + (y\, k_j)\sin\theta) \qquad \text{Eq. (1)}$$

where $a_j$ controls the curvature, $(h_j, k_j)$ is the peak of the parabola and $\theta_j$ is the angle of rotation relative to the x-axis.

In performing the preceding edge detection step, Wildes et al. bias the derivatives in the horizontal direction for detecting the eyelids, and in the vertical direction for detecting the outer circular boundary of the iris. The motivation for this is that the eyelids are usually horizontally aligned, and also the eyelid edge map will corrupt the circular iris boundary edge map if using all gradient data [12]. Taking only the vertical gradients for locating the iris boundary will reduce influence of the eyelids when performing circular Hough transform, and not all of the edge pixels defining the circle are required for successful localization. Not only does this make circle localization more accurate, it also makes it more efficient, since there are less edge points to cast votes in the Hough space.

## 2.2 Normalization

Once the iris region is successfully segmented from an eye image, the next stage is to transform the iris region so that it has fixed dimensions in order to allow comparisons. The normalization process will produce iris regions, which have the same constant dimensions, so that two photographs of the same iris under different conditions will have characteristic features at the same spatial location. We will be using Daugman's Rubber Sheet Model for normalization.

### 2.2.1 Daugman's Rubber Sheet Model

Daugman suggested normal Cartesian to polar transformation that maps each pixel in the iris area into a pair of polar coordinates $(r, \theta)$, where r and $\theta$ are on the intervals of [0 1] and [0 2$\pi$] [2].

This unwrapping can be formulated as

$$I(x(r, \theta), y(r, \theta)) \rightarrow I(r, \theta)$$

Such that

$$x(r, \theta) \rightarrow (1-r)\, x_p(\theta) + r\, x(\theta)$$

$$y(r, \theta) \rightarrow (1-r)\, y_p(\theta) + r\, y(\theta)$$

where $I(x, y)$, $(x, y)$, $(r, \theta)$, $(x_p, y_p)$, $(x_i, y_i)$ represent the iris region, Cartesian coordinates, polar coordinates, coordinates of the pupil and iris boundaries along $\theta$ direction respectively. Thus this representation often called as rubber sheet model.

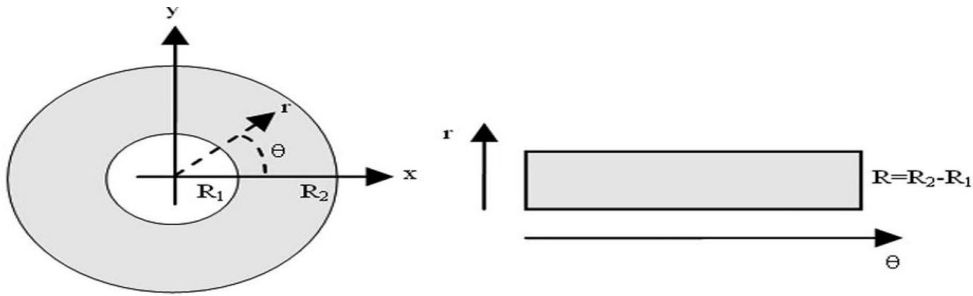

FIGURE 3: Daugman's rubber sheet model

The coordinate transformation process produces a 448 × 96 biometric pattern for each meaning ROI: 448 is the No. of the chosen radial samples (to avoid data loss in the round angle), while 96 pixels are the highest difference between iris and pupil radius in the iris images. In order to achieve invariance with regards to roto-translation and scaling distortion, the *r* polar coordinate is normalized in the [0, 1] range. For each Cartesian point of the ROI, image is assigned a polar coordinates pair *(r, θ)*, with $r \in [R1, R2]$ and $\theta \in [0, 2\pi]$, where *R1* is the pupil radius and *R2* is the iris radius, as shown in figure 3. The result of the normalized image is shown in figure 4.

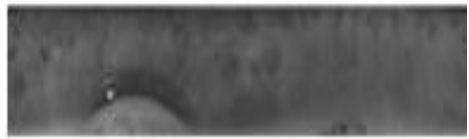

FIGURE 4: Normalized iris image

## 3. Feature Extraction

In order to provide accurate recognition of individuals, the most discriminating information present in an iris pattern must be extracted. Only the significant features of the iris must be encoded so that comparisons between templates can be made. Most iris recognition systems make use of a band pass decomposition of the iris image to create a biometric template [13].

The template that is generated in the feature encoding process will also need a corresponding matching metric, which gives a measure of similarity between two iris templates. This metric should give one range of values when comparing templates generated from the same eye, known as intra-class comparisons, and another range of values when comparing templates created from different irises, known as inter-class comparisons. These two cases should give distinct and separate values, so that a decision can be made with high confidence as to whether two templates are from the same iris, or from two different irises.

Feature extraction is a special form of dimensionality reduction. When the input data is too large to be processed and it is suspected to be notoriously redundant (much data, but not much information) then the input data will be transformed into a reduced representation set of features. For the purpose of feature extraction we will be Transforming into a reduced representation set of features is called feature extraction using the following three algorithms
- Zero crossing based 1-D wavelet

- Genetic algorithm
- Euler No.

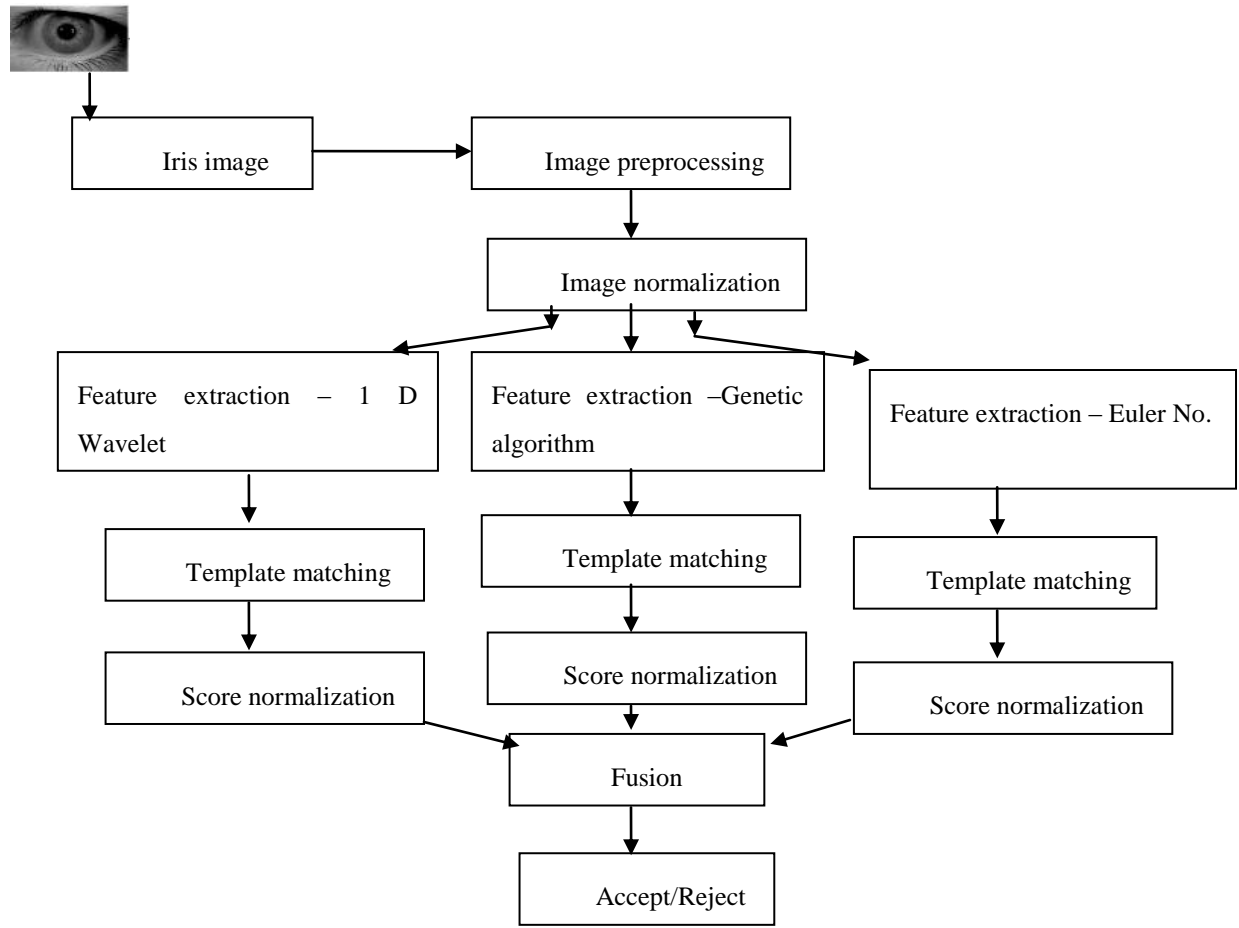

Fig. 5: Block diagram of proposed method

After extracting the features by this three algorithm, the template matching is done, for each individual, but the extracted features vary in size. Therefore a score normalization and fusion is required, as explained in section 5.

### 3.1 Zero crossing based 1-D wavelet

The mother wavelet is defined as the second derivative of a smoothing function $\theta(x)$. The zero crossings of dyadic scales of these filters are then used to encode features. The wavelet transform of a signal $f(x)$ at scale $s$ and position $x$ is given by

$$W_s f(x) = f * \left( s^2 \frac{d^2 \theta(x)}{dx^2} \right)(x) \qquad \text{Eq. (2)}$$

$$= s^2 \frac{d^2}{dx^2} (f * \theta_s)(x) \qquad \text{Eq. (3)}$$

Where,

$$\theta_s = (1/s)\theta(x/s)$$

$W_s f(x)$ is proportional to the second derivative of $f(x)$ smoothed by $\theta_s(x)$ and the zero crossings of the transform correspond to points of inflection in $f * \theta_s(x)$ [14]. The motivation for this technique is that zero-Crossings correspond to significant features with the iris region [15], [16], [17]. A correlation detection operator is defined as

$$G = \begin{array}{|c|c|c|} \hline -1 & 2 & -1 \\ \hline -1 & 2 & -1 \\ \hline -1 & 2 & -1 \\ \hline \end{array}$$

This detector is a 2D zero crossing detector. This operator is used for extracting iris feature by calculating convolution G and iris texture as shown in fig. 6.

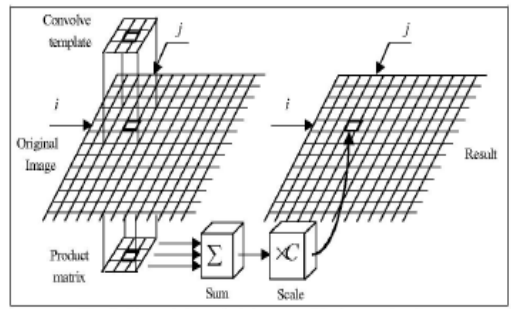

Fig 6. . Convolving sketch map of feature extraction

### 3.2 Eulers No.

Topological features, extracted using the Euler No. provide local information of iris patterns and are invariant to rotation, translation and scaling of the image. For a binary image, the Euler No. is defined as the difference between the No. of connected components and the No. of holes. Each pixel of unwrapped iris can be represented as an 8-bit binary vector {$b_7. b_6. b_5. b_4. b_3. b_2. b_1. b_0.$ }. These bits form eight planes with binary values as shown in fig 7. Four planes formed from the four most significant bits (MSBs) represent the structural information of the iris, and the remaining four planes represent the brightness information. The brightness information is random in nature and is not useful for comparing the structural features of two iris images [18]. The advantage here is role played by each individual bit is significant.

For comparing two iris images using the Euler code, a common mask is generated for both the iris images to be matched. The common mask is generated by performing a bitwise OR operation of the individual masks of the two iris images and is then applied to both the polar iris images. For each of the two iris images with a common mask, a 4-tuple Euler code is generated, which represent the Euler No. of the image corresponding to the four MSB planes.

We use mahalanobis distance to match the two Euler codes. The mahalanobis distance between two vectors is defined as $D(x, y) = \sqrt{(x-y)^t S^{-1} (x-y)}$, where x and y are the two Euler codes to be matched, and S is the positive-definite covariance matrix of x and y. if the Euler code has a large variance, it increases the false rejection rate. The mahalanobis distance ensures that the features having a high variance do not contribute to the distance. Applying the mahalanobis distance for comparison, thus, avoids the increase in the false reject rate.

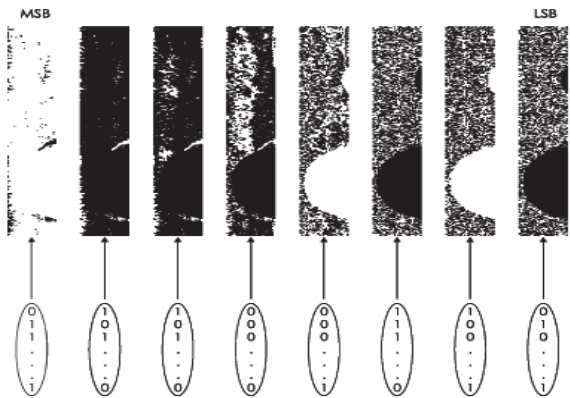

Fig. 7. Binary image corresponding to 8-bit planes of the masked polar image.

### *3.3 Genetic Algorithm*

GA select the prominent features based on the outcomes of the four features selection algorithms, namely the Entropy based approach, k-NN based method, T-statistics and the SVM-REF approach [19]. In order to obtain the most Selection algorithm subset from the different feature selection algorithms, a hybrid approach can be used. In order to choose the sets of feature selected by several features selection algorithms, four existing feature selection algorithm can be deployed, two filters (entropy-based, T-statistics) and two wrapper (SVM-REF, k-NNR) approaches to form feature pool. Apply each algorithm to the extracted features sequence and generate a ranking of those features. Give a ranking of features, we pick a No. of top ranked features from each algorithm and provide these top ranked features into the feature pool. The basic discussion of the four features selection algorithms can be found in [19]. Each individual represents a feature subset, and each individual in the population represents a candidate solution to the feature subset selection problem. It has the four basic steps:

| Initialization | : | Many individual solutions are randomly generated to form the population. |
|---|---|---|
| Selection | : | Individual solutions are selected through fitness based process where fitter solutions are selected. |
| Reproduction | : | Second generation population of solution are generated through genetic operators. They are: Crossover & Mutation. Process continues until a new |

| | |
|---|---|
| Termination : | population of solution is generated.<br>Process of reproduction is repeated until a termination condition has been reached. Conditions for termination are:<br>1. A solution is found that satisfies minimum Criteria.<br>2. Fixed No. of generations reached.<br>3. Allocated budget reached.<br>4. The highest ranking solution fitness is reached.<br>5. Manual inspections.<br>6. Combination of above. |

The above 4 steps are then applied to the features extracted through each of the gene from the selection pool and a fitness function is applied to it. Fitness function is defined as:

$$\text{Fitness} = W1.(1-RR) + W2.FAR + W3.FRR + W4.(\text{Feature size/Total No. of features}) \qquad \text{Eq. (4)}$$

where,  W1, W2, W3 and W4 are constant weighting parameters,
    RR – Recognition Rate,
    FAR – False Accept Rate,
    FRR– False Reject Rate.

We use Roulette wheel selection to probabilistically select individuals from a population for latter breeding. The probability of selecting an individual is estimated as

$$P(ind_i) = F(ind_i) / \Sigma^p_{i=1} F(ind_i) \qquad \text{Eq. (5)}$$

The probability that an individual will be selected is proportional to its own fitness and is inversely proportional to the fitness of the other competing hypothesis in the current population. Here, we use single point crossover, and each individual has a probability, $P_n$ to mutate. The No. of n bits is randomly selected to be flipped in every mutation stage.

 The advantage of using genetic algorithm is, we have n -No. of samples, out of which we select the samples which satisfies the fitness function and this process is repeated till we get the desired result.

**4.** *Template Matching*
 Template matching is the last process in the recognition of iris. This matching helps us to verify the authenticated person. The template matching compares the user template with the template from database using a matching metric. The matching metric compares similarity between two iris templates [20] [21].
  Template matching can be classified into 2 cases according to the matching metric. They are:
   1. Intra-class comparison
   2. Inter-class comparison
Intra-class comparison: When comparing templates are generated from the same iris. Inter-class comparison: When comparing templates are generated from different irises.
 The above two comparisons are then applied to the extracted features of the iris using Hamming distance. The bit obtained after the application decides the final output to be given. The bit 0 is obtained when intra-class comparison is performed and 1 bit is obtained when inter-class comparison is done.

## 5. Score Normalization And Score Fusion

Although it could be thought that learning based fusion are having better result than rule based fusion. Some examples have been reported in literature where the sum rule has outperformed other learning approaches [22].

Score fusion method consist of two stages. In the first stage, the output score of a single score $S_i$ is mapped onto a new score $S'_i$. this is referred as the normalization step or score mapping. Score normalization usually requires that several factors are known before the normalization is done, such as the range of the scores generated by the algorithm needs to be known. For eg. If algorithm X generates score between 0 and 100, a typical normalization step would be to divide the original score by 100. The second stage in the score fusion procedure is the fusion itself. Fusion usually takes as input one score per algorithm or modality and produces a single output score. There are many ways of fusing a set of scores to obtain a single score. The simplest fusion methods The simplest fusion methods include the average, the minimum, the maximum, and so on [23].

Last, an accept or reject decision is made on the test pattern xt using a threshold X, i.e.

$$\text{Result}(x_t) = \begin{cases} \text{accept, if output} \geq X \\ \text{Reject, if otherwise} \end{cases}$$

## 6. Conclusion & Future Work

The personal identification approaches using iris images are receiving increasing attention in the biometrics literature. Several methods have been presented in the literature and those based on the phase encoding of texture information are suggested to be the most promising. However, there has not been any attempt to combine these approaches to achieve further improvement in the performance. This paper presents a multialgorithmic fusion approach for iris recognition which combines the result obtained from scores of three algorithms namely Zero crossing based 1D wavelet, Genetic algorithm and Euler No., due to their advantages the combined approach would cover up the flaws in the process of feature extraction using single method and would increase the iris recognition performance. In future we will try to combine more efficient features extraction algorithm to get better efficiency and accuracy in iris recognition. Also we will try to fuse two or more modalities, like iris and fingerprint, to improve the performance over the unimodal systems.